\documentclass[10pt,twocolumn,letterpaper]{article}

\usepackage[pagenumbers]{cvpr} 
\usepackage{multirow}
\usepackage{xcolor}
\usepackage{colortbl}
\usepackage{utfsym}

\definecolor{cvprblue}{rgb}{0.21,0.49,0.74}
\usepackage[pagebackref,breaklinks,colorlinks,allcolors=cvprblue]{hyperref}

\def\paperID{32438} 
\def\confName{CVPR}
\def\confYear{2026}

\title{Spectral-Geometric Neural Fields for Pose-Free LiDAR View Synthesis}

\author{Yinuo Jiang\hspace{0.1in}
        Jun Cheng\hspace{0.1in} 
        Yiran Wang\footnotemark[1]\hspace{0.06in}\footnotemark[2]\hspace{0.1in}
        Cheng Cheng\footnotemark[1]\hspace{0.1in} \\
School of AIA, Huazhong University of Science and Technology\hspace{0.2in} 
\\
{\tt\small \{jiangyinuo,chengjun,wangyiran,c\_cheng\}@hust.edu.cn}\\
\vspace{-2mm}
}
\usepackage{etoolbox}
\preto{\appendix}{%
  \setcounter{figure}{0}%
  \setcounter{table}{0}%
}

\begin{document}
\maketitle
\renewcommand{\thefootnote}{\fnsymbol{footnote}} 
\footnotetext[1]{$\,$Corresponding author.}
\footnotetext[2]{$\,$Project leader.}
\begin{abstract}

Neural Radiance Fields (NeRF) have shown remarkable success in image novel view synthesis (NVS), inspiring extensions to LiDAR NVS. However, most methods heavily rely on accurate camera poses for scene reconstruction. The sparsity and textureless nature of LiDAR data also present distinct challenges, leading to geometric holes and discontinuous surfaces. To address these issues, we propose SG-NLF, a pose-free LiDAR NeRF framework that integrates spectral information with geometric consistency. Specifically, we design a hybrid representation based on spectral priors to reconstruct smooth geometry. For pose optimization, we construct a confidence-aware graph based on feature compatibility to achieve global alignment. In addition, an adversarial learning strategy is introduced to enforce cross-frame consistency, thereby enhancing reconstruction quality. Comprehensive experiments demonstrate the effectiveness of our framework, especially in challenging low-frequency scenarios. Compared to previous state-of-the-art methods, SG-NLF improves reconstruction quality and pose accuracy by over 35.8\% and 68.8\%. Our work can provide a novel perspective for LiDAR view synthesis.

\end{abstract}    
\section{Introduction}
\label{sec:Introduction}
Novel View Synthesis (NVS) is a vital task for 3D perception, with broad applications in scene understanding~\cite{li2025ch3depth,nvdsplus,li2024self,wang2022less}, autonomous driving~\cite{wang2025tacodepth,xiao2025densely}, and robotics~\cite{irshad2024neural,sethi2009validation}. The goal of NVS is to generate novel views from one or more given source views~\cite{li2022symmnerf}. While most existing works focus on synthesizing camera views, LiDAR sensors play an important role in autonomous systems for accurate perception. Extending NVS to LiDAR (\textit{i.e.}, LiDAR NVS) can significantly broaden perceptual field~\cite{zheng2024lidar4d,yu2025stgc} and enhance robustness of autonomous driving systems~\cite{wang2023neural,li2023diffusion,wang2021knowledge}.

Despite remarkable progress in image-based NVS, LiDAR NVS presents distinct challenges due to the inherent sparsity of LiDAR point clouds and their lack of texture information~\cite{li2025aeromamba,jiang2024gtinet,jiang2024robust,jiang2025registration}. Traditional LiDAR simulation approaches~\cite{dosovitskiy2017carla,koenig2004design,shah2017airsim} reconstruct explicit surfaces from aggregated point clouds and perform physically-based ray casting to render novel LiDAR views. However, such simulation methods struggle to accurately model the intensity or ray-drop characteristics of actual LiDAR points~\cite{yu2025stgc,huang2023neural}.

\begin{figure}
    \centering
    \includegraphics[trim=1cm 2cm 0cm 0cm, width=1\linewidth]{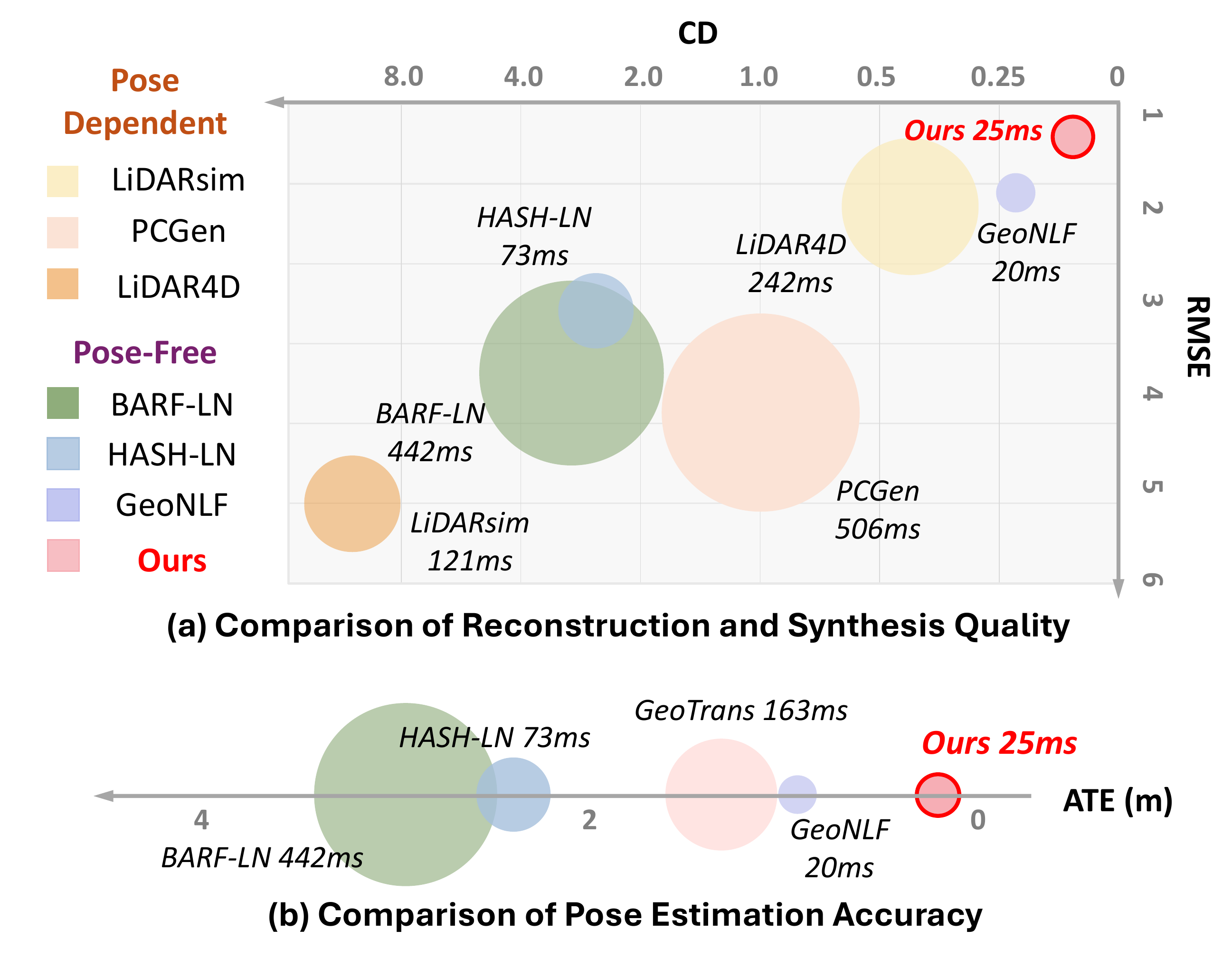}
    \vspace{-8pt}
    \caption{\textbf{(a) Reconstruction and synthesis quality.} The X-axis shows Chamfer Distance (CD) for point cloud reconstruction on the low-frequency KITTI-360 dataset~\cite{xue2024geonlf,liao2022kitti}. The Y-axis shows RMSE for range depth and intensity synthesis. Circle area represents inference time. Lower CD and RMSE mean better reconstruction and synthesis. Our framework outperforms prior work by large margins. \textbf{(b) Pose accuracy.} The axis represents Absolute Trajectory Error (ATE) for pose estimation. Our SG-NLF demonstrates higher accuracy compared to existing NeRF-based models.}
\vspace{-8pt}
    \label{fig:fig1}
\end{figure}

\begin{figure}
    \centering
    \includegraphics[trim=0.2cm 0cm 0cm 0cm, width=1\linewidth]{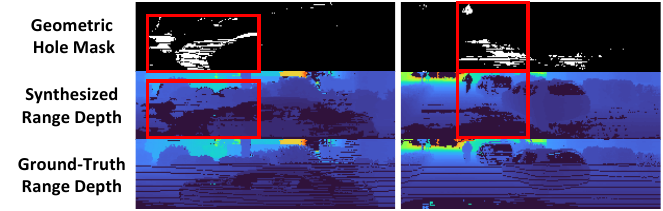}
    \vspace{-16pt}
\caption{\textbf{Geometric Inconsistency~\cite{tao2024lidar,huang2023neural,xue2024geonlf}.} The geometric hole mask is generated by comparing rendered opacity with ground truth LiDAR measurements. White regions in rectangular boxes (first row) highlight areas where these methods fail to reconstruct geometry. Comparisons between second and third rows further show these holes lead to poor-quality synthesized views.}

\vspace{-14pt}
    \label{fig:hole}
\end{figure}

Neural Radiance Fields (NeRF)~\cite{mildenhall2021nerf} provide a promising alternative by implicitly reconstructing scenes in continuous representation space through volume rendering. Recent advances have successfully extended NeRF to LiDAR novel view synthesis~\cite{zhang2024nerf,yang2023unisim,tao2024lidar,huang2023neural,zheng2024lidar4d,xue2024geonlf,yu2025stgc}, demonstrating superior performance over traditional methods~\cite{dosovitskiy2017carla,koenig2004design,shah2017airsim}. However, existing methods still face two critical challenges. First, as shown in Fig.~\ref{fig:fig1}, prior arts~\cite{zheng2024lidar4d,yu2025stgc,tao2024lidar,huang2023neural,yang2023unisim,zhang2024nerf} rely on accurate camera poses, which are often impractical to obtain~\cite{heo2023robust}. Recent pose-free methods like GeoNLF~\cite{xue2024geonlf} attempt to resolve this by simultaneous multi-view registration and neural reconstruction. However, the dependence on pairwise alignments makes it difficult to ensure global pose accuracy. Second, both pose-dependent and pose-free frameworks~\cite{tao2024lidar,huang2023neural,xue2024geonlf,zheng2024lidar4d,yu2025stgc} typically employ geometric interpolation (\textit{e.g.}, multi-resolution hash encoding~\cite{muller2022instant}) for neural field rendering. Due to the sparsity and irregularity of LiDAR data, such interpolated features often struggle to reconstruct continuous surfaces, leading to geometric inconsistency in textureless regions (shown in Fig.~\ref{fig:hole}). These challenges are further exacerbated in low-frequency LiDAR sequences, where large inter-frame motion and reduced overlap undermine multi-view consistency~\cite{yu2025stgc,xue2024geonlf}.

To address the aforementioned challenges, we propose \textbf{SG-NLF}, a spectral-geometric neural LiDAR fields framework for novel view synthesis and pose estimation. Specifically, we design a hybrid representation that integrates spectral embeddings with geometric encoding to reconstruct smooth and consistent geometry. For pose optimization, we construct a confidence-aware graph and establish geometric constraints through feature compatibility, achieving global pose estimation. We further introduce an adversarial learning strategy that enforces cross-frame consistency and reconstruction quality through discriminative supervision. As shown in Fig.~\ref{fig:fig1}, our SG-NLF can simultaneously achieve high-quality LiDAR view synthesis and accurate pose estimation, outperforming previous state-of-the-art methods.

We evaluate our SG-NLF on the KITTI-360~\cite{liao2022kitti} and nuScenes~\cite{caesar2020nuScenes} datasets, focusing on low-frequency sequences that span hundreds of meters~\cite{xue2024geonlf}. Compared to the leading method GeoNLF~\cite{xue2024geonlf}, SG-NLF reduces Chamfer Distance by \textbf{35.8\%} and Absolute Trajectory Error by \textbf{68.8\%} on nuScenes, with consistently significant improvements across all metrics. Additionally, on standard-frequency sequences with complex scenes~\cite{tao2024lidar}, SG-NLF continues to outperform prior arts, demonstrating the generalization. Our contributions are summarized as follows:

\begin{itemize}
\item We propose SG-NLF, a novel framework that integrates spectral-geometric information for simultaneous high-quality view synthesis and accurate pose estimation.
\item We design a hybrid representation that combines spectral embeddings for geometry-consistent reconstruction and a confidence-aware graph for global pose optimization.
\item Comprehensive experiments prove our state-of-the-art performance even in challenging low-frequency scenes.
\end{itemize}

\section{Related Work}
\label{sec:Related Work}

\noindent \textbf{LiDAR Simulation.} Traditional LiDAR simulators~\cite{dosovitskiy2017carla,koenig2004design} generate point clouds through physics-based ray casting in virtual environments, showing a significant domain gap compared to real-world data~\cite{yu2025stgc,zheng2024lidar4d}. Reconstruct-then-simulate approaches~\cite{manivasagam2020lidarsim,li2022pcgen} attempt to bridge this gap by combing realistic LiDAR data. LiDARsim~\cite{manivasagam2020lidarsim} performs reconstruction using mesh surfel representation, while PCGen~\cite{li2022pcgen} reconstructs directly from point clouds and employs rasterization for rendering. However, these methods still rely on explicit surface reconstruction, making them difficult to accurately model the intensity or ray drop characteristics of actual LiDAR points~\cite{xue2024geonlf,tao2024lidar,zheng2024lidar4d,yu2025stgc}.


\noindent \textbf{NeRF for LiDAR View Synthesis.} Neural Radiance Fields (NeRF)~\cite{mildenhall2021nerf} have shown remarkable progress in image novel view synthesis~\cite{li2022symmnerf,huang2023neural}, with many variants built on voxel grids~\cite{fridovich2022plenoxels,liu2020neural} and multi-resolution hash encodings~\cite{muller2022instant}. Recent studies have extended NeRF to LiDAR novel view synthesis, demonstrating performance that surpasses traditional simulation approaches~\cite{manivasagam2020lidarsim,li2022pcgen}. Early methods such as NeRF-LiDAR~\cite{zhang2024nerf} leverage both RGB images and LiDAR point clouds for scene reconstruction. However, such multi-modal data are not always available in practical applications~\cite{yu2025stgc}. LiDAR-NeRF~\cite{tao2024lidar} and NFL~\cite{huang2023neural} first introduce differentiable frameworks for LiDAR NVS, which synthesize depth, intensity, and ray-drop probability without relying on RGB inputs. Subsequent works like LiDAR4D~\cite{zheng2024lidar4d} and STGC~\cite{yu2025stgc} further extend the framework to dynamic scenes by incorporating temporal priors. A commonality across these methods is the sole reliance on geometric interpolation for neural field rendering. Due to the inherent sparsity and irregularity of LiDAR data, such representation often struggles to reconstruct continuous surfaces, leading to geometric inconsistency in textureless regions. To overcome this limitation, we propose a novel hybrid representation that combines spectral embeddings to reconstruct smooth and consistent geometry.

\noindent \textbf{Pose-free NeRF.}  
In addition to the reconstruction challenges, most existing LiDAR NVS methods~\cite{zhang2024nerf,tao2024lidar,huang2023neural,zheng2024lidar4d,yu2025stgc} exhibit a strong dependence on known accurate poses for rendering, which are often difficult to acquire in real-world scenarios~\cite{lin2021barf}. Inspired by the image-based pose-free methods like BARF~\cite{lin2021barf} and HASH~\cite{heo2023robust}, GeoNLF~\cite{xue2024geonlf} introduces pose-free neural LiDAR fields to optimize  multi-view registration and neural reconstruction simultaneously. Despite this progress, the pairwise alignment constraints limit the global trajectory accuracy. In this paper, we propose a confidence-aware pose graph based on spectral-geometric feature compatibility for global optimization.

\begin{figure*}[!t]
    \centering
    \includegraphics[trim=0cm 0cm 0cm 0cm, clip, width=1\linewidth]{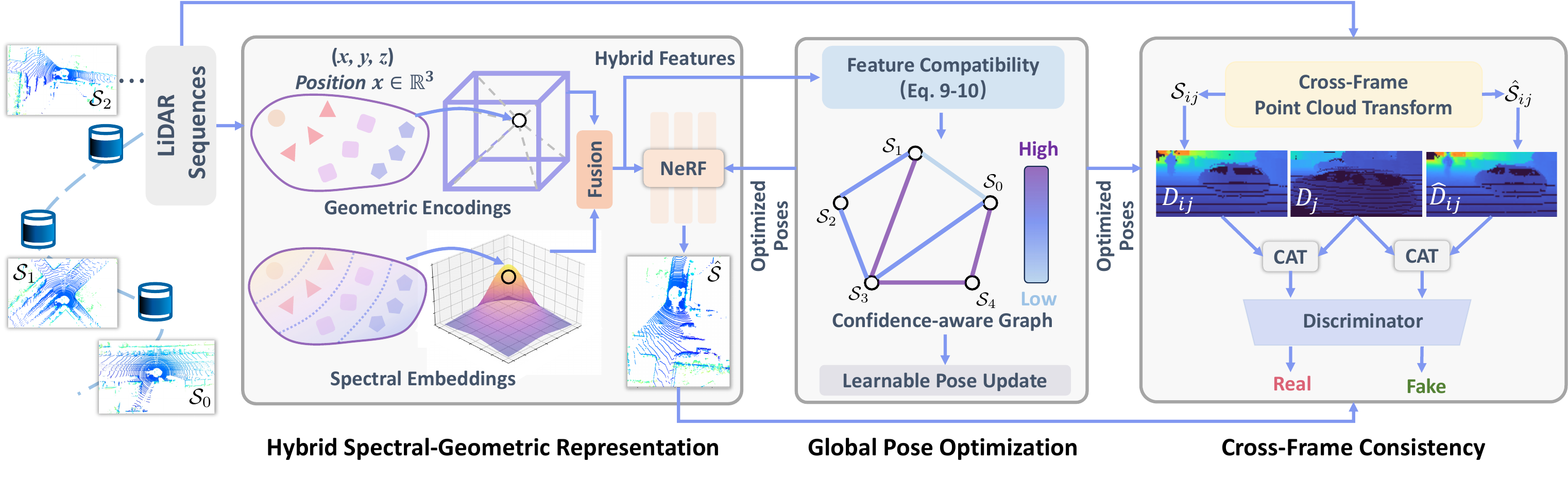}
    \vspace{-20pt}
\caption{\textbf{Overview of the SG-NLF.} Given multi-view LiDAR sequences $\{\mathcal{S}_i\}_{i=0}^{N}$, SG-NLF employs a hybrid representation that combines discrete geometric encoding with continuous spectral embedding for effective reconstruction. The hybrid features are utilized to construct a confidence-aware graph for global pose optimization. The optimized poses and hybrid features are fed into NeRF to synthesize novel views $\hat{\mathcal{S}}$.  To further enhance cross-frame consistency, SG-NLF distinguishes between the real depth maps (obtained from ground truth point clouds $\mathcal{S}_{ij}$ that are transformed into cross-frame coordinates) and fake depth maps (obtained from synthesized and transformed $\hat{\mathcal{S}}_{ij}$).}

\vspace{-15pt}
    \label{fig:overview}
\end{figure*} 
\section{Methodology}

We propose SG-NLF, a spectral-geometric neural LiDAR fields framework for pose-free view synthesis. In this section, we first introduce the problem formulation and preliminary for LiDAR NeRF with pose optimization, followed by a detailed description of our proposed SG-NLF method.

\subsection{Overview}
\noindent \textbf{Problem Formulation.} Given LiDAR point cloud sequences $\{\mathcal{S}_i\}_{i=0}^{N}$ captured in a large-scale outdoor scene, our goal is to recover their global poses $\{\mathbf{T}_i\}_{i=0}^{N}$ and jointly reconstruct the scene as a continuous implicit representation. The reconstructed representation can synthesize realistic LiDAR point clouds from an arbitrary new viewpoint.
.

\noindent \textbf{Preliminary for NeRF with Pose Optimization.} Following prior LiDAR NeRF frameworks~\cite{xue2024geonlf,zheng2024lidar4d,yu2025stgc,tao2024lidar}, we project LiDAR point clouds into range images, where each laser beam is modeled as a ray $\mathbf{r}(t) = \mathbf{o} + t\mathbf{d}$ originating from the sensor center $\mathbf{o}$ with direction $\mathbf{d}$. The neural function is queried at 3D points sampled along $\mathbf{r}$ to predict their depth value $z$ and volume density $\sigma$. Volume rendering is employed to compute the pixel depth $\hat{\mathcal{D}}(\mathbf{r})$:
\begin{equation}
\small
\begin{aligned}
\hat{\mathcal{D}}(\mathbf{r})=\sum_{i=1}^N T_i\left(1-e^{-\sigma_i \delta_i}\right) z_i, \,
T_i=\exp(-\sum_{j=1}^{i-1} \sigma_j \delta_j)\,,
\end{aligned}
\end{equation}
where $\delta$ is the distance between samples and $T$ represents the accumulated transmittance. Intensity and ray-drop probability are predicted separately in the same way~\cite{tao2024lidar,xue2024geonlf}.

While original NeRFs require precise camera poses for rendering, pose-free methods treat them as learnable parameters~\cite{xue2024geonlf,lin2021barf}. We follow prior work~\cite{xue2024geonlf,lin2022parallel} to optimize poses in their corresponding Lie algebra space. Each pose is parameterized by a 6D vector $\boldsymbol{\xi}=[\boldsymbol{\rho}, \boldsymbol{\phi}]^{\top}  \in \mathbb{R}^6$ and updated by computing a small increment $\Delta\boldsymbol{\xi}$ in the algebra space, such that $\boldsymbol{\xi}' = \boldsymbol{\xi} + \Delta\boldsymbol{\xi}$. The transformation is recovered from the vector $\boldsymbol{\xi}$ via the exponential mapping:
\begin{equation}
\mathbf{T} = \exp(\boldsymbol{\xi}^{\wedge})=\sum_{n=0}^{\infty} \frac{1}{n!}\left(\boldsymbol{\xi}^{\wedge}\right)^n = \begin{bmatrix}
\boldsymbol{R} & \boldsymbol{J}\boldsymbol{\rho} \\
\mathbf{0}^\top & 1
\end{bmatrix},
\end{equation}
where $\boldsymbol{R}=\exp(\boldsymbol{\phi}^{\wedge})$ and $\boldsymbol{J}=\sum_{n=0}^{\infty} \frac{1}{(n+1)!}(\boldsymbol{\phi}^{\wedge})^n$. The Jacobian $\boldsymbol{J}$ couples rotation and translation components in the Lie algebra update, which may lead to non-intuitive optimization trajectories when using momentum-based optimizers. We follow prior arts~\cite{xue2024geonlf,lin2022parallel} to omit the Jacobian $\boldsymbol{J}$ from the translation term for more stable convergence.


\noindent \textbf{Overview of SG-NLF Framework.} Fig.~\ref{fig:overview} illustrates an overview of the proposed SG-NLF framework. Existing methods~\cite{tao2024lidar,huang2023neural} rely solely on interpolating 3D positions to extract features and render neural fields, which often results in geometric inconsistency. In contrast, our SG-NLF introduces a hybrid representation that integrates discrete geometric encodings with continuous spectral embeddings, enabling more effective and robust reconstruction. Regarding pose optimization, compared to previous pose-free approaches with pairwise alignment constraints~\cite{xue2024geonlf}, SG-NLF constructs a confidence-aware graph based on hybrid feature compatibility, achieving global pose optimization. The optimized poses, together with the hybrid features, are fed into NeRF to synthesize LiDAR views $\hat{\mathcal{S}}$. To further improve cross-frame consistency, SG-NLF employs a discriminative supervision mechanism that distinguishes between the real depth maps (obtained from the ground truth point clouds $\mathcal{S}_{ij}$ that are transformed into the cross-frame coordinate system) and the fake depth maps (generated from synthesized and transformed point clouds $\hat{\mathcal{S}}_{ij}$).

\subsection{Hybrid Spectral-Geometric Representation}
Our representation builds upon multi-resolution hash grid encoding~\cite{muller2022instant} to extract geometric features $\boldsymbol{f}_\text{geo}(\mathbf{x})$, capturing local structures and high-frequency details~\cite{huang2023neural,zheng2024lidar4d}. 
However, the inherent sparsity of LiDAR point clouds means that interpolation-based features alone are insufficient, as they struggle to reconstruct complete and coherent surfaces in unobserved regions due to the lack of global structural priors~\cite{jiang2023neural}. To overcome this limitation, we augment the geometric representation with spectral embeddings, which exhibit intrinsic isometry invariance and are aware of the underlying surface geometry~\cite{jiang2023neural,williamson2025neural,levy2010spectral}.

While spectral embeddings are traditionally obtained via an inefficient and non-differentiable offline eigen-decomposition of the Laplace-Beltrami operator (LBO)~\cite{reuter2010hierarchical,levy2010spectral}, we draw inspiration from recent neural spectral approaches~\cite{jiang2023neural,williamson2025neural}, which optimize a set of MLPs and approximate the first $K$ LBO eigenfunctions in a differentiable manner to obtain spectral embeddings. We construct the underlying scene geometry following SNS~\cite{williamson2025neural} and represent the scalar fields using MLPs. The spectral embeddings are then defined as $\boldsymbol{f}_\text{spe}(\mathbf{x}) = [\Psi_0(\mathbf{x}), \Psi_1(\mathbf{x}), \dots, \Psi_K(\mathbf{x})]^{\top}$, where $\Psi_0(\mathbf{x})$ is the constant eigenfunction corresponding to the zero eigenvalue. To ensure our learned embeddings exhibit the intrinsic spectral properties, we optimize for the first $K$ non-constant eigenfunctions by minimizing the Rayleigh quotient. Our implementation of the Laplace-Beltrami operator follows the mean curvature formulation~\cite{williamson2025neural}. To ensure uniform sampling on the implicit surface, we adopt the rejection sampling strategy from SNS~\cite{williamson2025neural} and sample $M$ points $\{\hat{\mathbf{x}}_i\}_{i=1}^M$ from the surface, where $M$ is the number of Monte Carlo samples used for discrete approximation. For each sampled point $\hat{\mathbf{x}}_i$, we compute the local area element $dA_i$ using the First Fundamental Form:
\begin{equation}
dA_i = \sqrt{E_i \cdot G_i - F_i^2}\,,
\end{equation} 
where $E_i = \|\mathbf{t}_{i,1}\|^2$, $G_i = \|\mathbf{t}_{i,2}\|^2$, and $F_i = \langle \mathbf{t}_{i,1}, \mathbf{t}_{i,2} \rangle$ are coefficients of the First Fundamental Form, with $\mathbf{t}_{i,1}$ and $\mathbf{t}_{i,2}$ being tangent plane basis vectors computed via automatic differentiation of the implicit surface mapping~\cite{williamson2025neural}. The surface gradient of the eigenfunction is computed by projecting its 3D spatial gradient onto the tangent plane:
\begin{equation} 
\nabla_{\Sigma} \Psi(\hat{\mathbf{x}}_i) = \nabla \Psi(\hat{\mathbf{x}}_i) - \langle \nabla \Psi(\hat{\mathbf{x}}_i), \mathbf{n}(\hat{\mathbf{x}}_i) \rangle \mathbf{n}(\hat{\mathbf{x}}_i)\,, 
\end{equation}
where $\mathbf{n}(\hat{\mathbf{x}}_i)$ denotes the unit normal vector, computed as the normalized cross product of tangent plane basis vectors: $\mathbf{n}(\hat{\mathbf{x}}_i) = \frac{\mathbf{t}_{i,1} \times \mathbf{t}_{i,2}}{\|\mathbf{t}_{i,1} \times \mathbf{t}_{i,2}\|}$. Using these robust geometric definitions, the discrete Rayleigh quotient~\cite{williamson2025neural} is formulated as:
\begin{equation}
\mathcal{R}_{\Sigma}(\Psi_i) =
\frac{\sum_{j=1}^M \|\nabla_{\Sigma} \Psi_i(\hat{\mathbf{x}}_j)\|^2 dA_j}
{\sum_{j=1}^M \Psi_i^2(\hat{\mathbf{x}}_j) dA_j} \,.
\end{equation}

To ensure that the learned eigenfunctions form a valid orthonormal basis on the surface, we impose both orthogonality and normalization constraints~\cite{williamson2025neural}. The orthogonality loss enforces distinct spectral components:
\begin{equation}
\mathcal{L}_{\text{ortho}} = \sum_{i=1}^K \sum_{\substack{m=1 \\ m \neq i}}^K \left( \sum_{j=1}^M \Psi_i(\hat{\mathbf{x}}_j) \Psi_m(\hat{\mathbf{x}}_j) dA_j \right)^2\,,
\end{equation}
while the normalization loss maintains unit $L^2$-norm for each eigenfunction:
\begin{equation}
\vspace{-2pt}
\mathcal{L}_{\text{norm}} = \sum_{i=1}^K \left( \sum_{j=1}^M \Psi_i^2(\hat{\mathbf{x}}_j) dA_j - 1 \right)^2\,.
\end{equation}

The overall spectral loss is formulated as:
\begin{equation}
\vspace{-6pt}
\mathcal{L}_\text{spe} = \sum_{i=1}^{K} \mathcal{R}_{\Sigma}(\Psi_i) + \lambda_n \mathcal{L}_{\text{norm}} + \lambda_o \mathcal{L}_{\text{ortho}}\,,
\vspace{-0.5pt}
\end{equation}
where $\lambda_n$ and $\lambda_o$ are balancing coefficients. The geometric encodings and spectral embeddings are progressively fused during training to form a hybrid representation $\boldsymbol{f}_\text{hyb}(\mathbf{x})$, which is subsequently used by the rendering network to predict depth, intensity, and ray-drop probability.
\subsection{Global Pose Optimization}
We introduce a pose graph that is dynamically constructed through spectral-geometric feature matching. The graph is defined as $\mathcal{G} = (\mathcal{V}, \mathcal{E})$. Each vertex $v_i \in \mathcal{V}$ represents a LiDAR point cloud $\mathcal{S}_i$ with its global pose $\mathbf{T}_i$, while edge $(i, j) \in \mathcal{E}$ represents constraints between $\mathcal{S}_i$ and $\mathcal{S}_j$. Compared to approaches~\cite{xue2024geonlf} that rely solely on enforcing constraints between temporal neighbors, our edge set $\mathcal{E}$ incorporates both sequential edges and non-adjacent edges with high feature compatibility, thereby jointly enforcing inter-frame alignment and global trajectory accuracy.

To identify reliable non-adjacent edges, we first leverage our fused spectral-geometric features $\boldsymbol{f}_\text{hyb}(\mathbf{x})$ to establish point-wise correspondences. For the $m$-th point in $\mathcal{S}_i$, denoted as $\mathbf{x}^i_m$, its corresponding feature is defined as $\boldsymbol{f}^i_m = \boldsymbol{f}_\text{hyb}(\mathbf{x}^i_m)$. Similarly, for the $n$-th point in $\mathcal{S}_j$, the feature is $\boldsymbol{f}^j_n = \boldsymbol{f}_\text{hyb}(\mathbf{x}^j_n)$. We adopt a coarse-to-fine Mutual Nearest Neighbor (MNN) strategy to build reliable correspondences. For a candidate scan pair $(\mathcal{S}_i, \mathcal{S}_j)$, the coarse correspondence set is constructed as:
\begin{equation}
\vspace{-2pt}
\small
\begin{aligned}
\mathcal{M}_{c}^{ij} = \Bigg\{ (m, n) \,\Bigg|\, & \|\boldsymbol{f}^i_m - \boldsymbol{f}^j_n\|_2 = \min_{n'} \|\boldsymbol{f}^i_m - \boldsymbol{f}^j_{n'}\|_2 \land \\&\|\boldsymbol{f}^j_n - \boldsymbol{f}^i_m\|_2 = \min_{m'} \|\boldsymbol{f}^j_n - \boldsymbol{f}^i_{m'}\|_2 \Bigg\}\,,
\end{aligned}
\end{equation}
where $ \land$ denotes the logical AND operator. Then, a refined correspondence set $\mathcal{M}_{f}^{ij}$ is generated by reapplying MNN matching in $\mathcal{M}_{c}^{ij}$. We calculate the mean cosine similarity of feature pairs in $\mathcal{M}_{f}^{ij}$ as the edge compatibility score:
\begin{equation}
E^{ij} = \frac{1}{|\mathcal{M}_{f}^{ij}|} \sum_{(m,n) \in \mathcal{M}_{f}^{ij}} \frac{ \boldsymbol{f}^i_m \cdot \boldsymbol{f}^j_n }{ \|\boldsymbol{f}^i_m\|_2 \|\boldsymbol{f}^j_n\|_2 }\,,
\end{equation}
which serves as the basis for edge inclusion. An edge $(i, j)$ is added to $\mathcal{E}$ if $E^{ij}$ exceeds an adaptive threshold and threshold is progressively increased during training to enforce stricter edge selection. To further mitigate the influence of inaccurate edges, each edge is weighted by the correspondence spatial consistency. For $(i, j) \in \mathcal{E}$, we follow prior arts~\cite{bai2021pointdsc,qin2022geometric} to compute the spatial consistency score $P_{mn}$ for point correspondence $(m,n) \in \mathcal{M}_{f}^{ij}$:
\begin{equation}
\small
P_{mn} = \frac{1}{N_{m}} \sum_{\substack{(m',n') \in \mathcal{M}_f^{ij}, \\ m \neq m'}} \mathbb{I}\big( | d_{mm'}^i - d_{nn'}^j | < \tau_d \big)\,,
\end{equation}
where $\mathcal{N}_m$ is the count of correspondence pairs in $\mathcal{M}_f^{ij}$ excluding $(m,n)$. $d_{mm'}^i = \|\mathbf{x}_m^i - \mathbf{x}_{m'}^i\|_2$ and $d_{nn'}^j = \|\mathbf{x}_n^j - \mathbf{x}_{n'}^j\|_2$ represent the length distances between correspondence pairs. $\tau_d$ is a distance preservation threshold. We compute the edge weight for each edge based on the average spatial consistency:
\begin{equation}
\alpha^{ij} = \frac{1}{|\mathcal{M}_{f}^{ij}|} \sum_{(m,n)\in \mathcal{M}_{f}^{ij}} P_{mn}\,.
\end{equation}

The pose graph loss is formulated as the weighted Chamfer Distance loss:
\begin{equation}
\mathcal{L}_\text{graph} = \sum_{(i,j)\in \mathcal{E}} \alpha^{ij} \cdot \mathcal{L}_{\text{cd}}^{ij}\,,
\vspace{-6pt}
\end{equation}
where $\mathcal{L}_{\text{cd}}^{ij}$ is Chamfer Distance~\cite{fan2017point} computed following prior arts~\cite{xue2024geonlf}. Refer to supplement for more formulations.

\begin{table*}
\caption{\textbf{Comparisons with state-of-the-art methods on KITTI-360~\cite{liao2022kitti} dataset with a low-frequency setting~\cite{xue2024geonlf}.} We compare our method with pose-dependent~\cite{manivasagam2020lidarsim,li2022pcgen,zheng2024lidar4d} and pose-free methods~\cite{lin2021barf,heo2023robust,xue2024geonlf,qin2022geometric}. For approaches not originally designed for LiDAR NVS~\cite{lin2021barf,heo2023robust,qin2022geometric}, we follow GeoNLF~\cite{xue2024geonlf} and adopt LiDAR-NeRF~\cite{tao2024lidar} for reconstruction. For pose-dependent methods, we employ ground-truth poses for synthesis. We color
the best results as \textcolor{red}{red} and the second-best as \textcolor{orange}{orange}.}
\vspace{-3pt}
    \centering
    \resizebox{\textwidth}{!}{
    \begin{tabular}{lccccccccccccccc}

        \hline
        \multirow{2}*{Method}  & \multicolumn{3}{c}{Point Cloud} & \multicolumn{6}{c}{Depth} & \multicolumn{5}{c}{Intensity}\\
        \cmidrule{2-3} \cmidrule{5-9} \cmidrule{11-15}
                            & CD$\downarrow$ &F-score$\uparrow$& &RMSE$\downarrow$ & MedAE$\downarrow$ & LPIPS$\downarrow$& SSIM$\uparrow$& PSNR$\uparrow$ && RMSE$\downarrow$ & MedAE$\downarrow$ & LPIPS$\downarrow$& SSIM$\uparrow$& PSNR$\uparrow$\\
        \hline
        LiDARsim~\cite{manivasagam2020lidarsim}                 & 11.0426  &0.5975 & & 10.1994  & 1.3881 & 0.5588 & 0.3888  & 17.9423 && 0.2089   &  0.1254  & 0.6463  & 0.0824  & 13.6127\\
        PCGen~\cite{li2022pcgen}                 & 1.0356  &0.7862 & & 7.5672  & 0.7066 & 0.5333 & 0.3744  & 20.6219 && 0.2139    & 0.1063  & 0.5930 & 0.1070  & 13.4243\\
        
        BARF-LN~\cite{lin2021barf,tao2024lidar}         & 3.1001  &0.6156 & & 7.5767  & 2.0583 & 0.5779 & 0.2834  & 22.5759& & 0.2121    & 0.1575  & 0.7121  & 0.1468  & 11.9778\\
        HASH-LN~\cite{heo2023robust,tao2024lidar}   & 2.6913 & 0.6082 & & 6.3005 & 2.1686  & 0.5176 & 0.3752  & 22.6196& & 0.2404    & 0.1502   & 0.6508  & 0.1602 & 12.9286\\
        GeoTrans-LN~\cite{qin2022geometric,tao2024lidar}           & 0.5753   & 0.8116 & & 4.4291  & 0.2023  & 0.3896  & 0.5330 & \cellcolor{orange!40}25.6137  && 0.2709    & 0.1589   & 0.5578  & 0.2578  & 12.9707\\
LiDAR4D~\cite{zheng2024lidar4d}              & 0.2760  &0.8843 & & 4.7303  & \cellcolor{orange!40}0.0785 & \cellcolor{orange!40}0.3368 & 0.6197  & 24.7282 & &\cellcolor{orange!40}0.1459    & \cellcolor{orange!40}0.0524  & \cellcolor{orange!40}0.3883& \cellcolor{orange!40} 0.3406    & \cellcolor{orange!40}16.9512\\
        GeoNLF~\cite{xue2024geonlf}                 & \cellcolor{orange!40}0.2363   & \cellcolor{orange!40}0.9178  && \cellcolor{orange!40}4.0293  & 0.1009  & 0.3900  & \cellcolor{orange!40}0.6272  & 25.2758 & & 0.1495     & 0.1525     & 0.5379    & 0.3165  & 16.5813\\
       SG-NLF(Ours)                  & \cellcolor{red!40}0.1695   & \cellcolor{red!40}0.9191 && \cellcolor{red!40}2.9514 & \cellcolor{red!40}0.0544 & \cellcolor{red!40}0.0701  & \cellcolor{red!40}0.9270  & \cellcolor{red!40}28.7068  & &\cellcolor{red!40}0.1089     & \cellcolor{red!40}0.0368   & \cellcolor{red!40}0.2026  & \cellcolor{red!40} 0.5751  & \cellcolor{red!40}19.2652\\
       \hline
    \end{tabular}}

    \label{tab:kitti}
\end{table*}

\begin{table*}
\caption{\textbf{Comparisons with state-of-the-art methods on nuScenes~\cite{caesar2020nuScenes} dataset with a low-frequency setting~\cite{xue2024geonlf}.} We color
the best results as \textcolor{red}{red} and the second-best as \textcolor{orange}{orange}. The notations are consistent with the KITTI-360~\cite{liao2022kitti} dataset in Table~\ref{tab:kitti} above.}
\vspace{-3pt}
    \centering
    \resizebox{\textwidth}{!}{
    \begin{tabular}{lccccccccccccccc}

        \hline
        \multirow{2}*{Method}  & \multicolumn{3}{c}{Point Cloud} & \multicolumn{6}{c}{Depth} & \multicolumn{5}{c}{Intensity}\\
        \cmidrule{2-3} \cmidrule{5-9} \cmidrule{11-15}
                            & CD$\downarrow$ &F-score$\uparrow$& &RMSE$\downarrow$ & MedAE$\downarrow$ & LPIPS$\downarrow$& SSIM$\uparrow$& PSNR$\uparrow$ && RMSE$\downarrow$ & MedAE$\downarrow$ & LPIPS$\downarrow$& SSIM$\uparrow$& PSNR$\uparrow$\\
        \hline
        LiDARsim~\cite{manivasagam2020lidarsim}               & 16.7623  &0.4308 & & 12.3483  & 1.9971 & 0.3125 & 0.3889  & 16.2553 && 0.0858    & 0.0355  & 0.1713  & 0.2897  & 21.3989\\
        PCGen~\cite{li2022pcgen}               & 2.2608  &0.6139  && 12.6586  &0.7055 &  0.2365 & 0.4389  & 12.6586& & 0.0865    & 0.0235  & 0.1530  & 0.3707  & 21.3643\\   
        BARF-LN~\cite{lin2021barf,tao2024lidar}               & 1.2695   & 0.7589 & & 8.2414   & 0.1123  & 0.1432  & 0.6856  & 20.8900 &  & 0.3920   & 0.0144    & 0.1023   & 0.6119 & 26.2330\\
        HASH-LN~\cite{heo2023robust,tao2024lidar}             & 0.9691   & 0.8011 & & 7.8353  & 0.0441 & 0.1190  & 0.6543  & 20.6244 & & 0.0459     & 0.0135   & 0.0954  & 0.6279  & 26.8870\\
        GeoTrans-LN~\cite{qin2022geometric,tao2024lidar}              & 4.1587    & 0.2993 &  & 10.7899   & 2.1529   & 0.1445   & 0.3671   & 17.5885 && 0.0679     & 0.0256  & 0.1149  & 0.3476  & 23.6211\\
        LiDAR4D~\cite{zheng2024lidar4d}               & 0.5668  & 0.7444 & & 11.1964  & 0.0847 & \cellcolor{orange!40}0.0685 & 0.5983  & 17.0920 && 0.0599    & 0.0180  & \cellcolor{orange!40}0.0639 & 0.5401  & 24.4753\\    
        GeoNLF~\cite{xue2024geonlf}              & \cellcolor{orange!40}0.2408   & \cellcolor{orange!40}0.8647   & &\cellcolor{orange!40}5.8208   & \cellcolor{orange!40}0.0281   & 0.0727  & \cellcolor{orange!40}0.7746   & \cellcolor{orange!40}22.9472  & &\cellcolor{orange!40}0.0378      & \cellcolor{orange!40}0.0100   & 0.0774   & \cellcolor{orange!40}0.7368  & \cellcolor{orange!40}28.6078\\
        SG-NLF(Ours)          & \cellcolor{red!40}0.1545  & \cellcolor{red!40}0.9097 & &\cellcolor{red!40}3.0706 & \cellcolor{red!40}0.0278 & \cellcolor{red!40}0.0191 & \cellcolor{red!40}0.9398 & \cellcolor{red!40}28.4094 &&\cellcolor{red!40}0.0299    & \cellcolor{red!40}0.0078   & \cellcolor{red!40}0.0349  & \cellcolor{red!40} 0.8679 & \cellcolor{red!40}30.4987\\
        \hline
    \end{tabular}}
    \vspace{-10pt}

    \label{tab:nus}
\end{table*}

\subsection{Cross-frame Consistency}
Most existing LiDAR NeRF frameworks~\cite{tao2024lidar,zheng2024lidar4d,xue2024geonlf} only rely on pixel-wise supervision of range images, which penalizes photometric errors within individual frames but overlooks structural information. To enhance cross-frame consistency, we introduce an adversarial learning strategy that evaluate quality of reconstructed geometry and accuracy of pose estimation through discriminative supervision.

For temporally adjacent frames $(i, j)$, we compute the estimated relative pose as \(\hat{\mathbf{T}}_{ij} = \hat{\mathbf{T}}_{j}^{-1} \hat{\mathbf{T}}_{i}\). This transformation is applied to the reconstructed point cloud \(\hat{\mathcal{S}}_i\), yielding the transformed point cloud \(\hat{\mathcal{S}}_{ij}\) in the coordinate system of frame $j$. We then render a depth map $\hat{D}_{ij}$ from $\hat{\mathcal{S}}_{ij}$ and concatenate $\hat{D}_{ij}$ with the ground truth depth map $D_j$ of frame $j$, forming \(\mathbf{I}_{\text{fake}} = [\hat{D}_{ij}, D_j]\). Correspondingly, the ground truth point cloud $\mathcal{S}_i$ is transformed using the ground truth relative pose $\mathbf{T}_{ij}$ and projected to produce $D_{ij}$. This is concatenated with $D_j$ to construct the real sample $\mathbf{I}_{\text{real}} = [D_{ij}, D_j]$. By incorporating paired depth maps from adjacent frames, the discriminator is able to assess not only the quality of per-frame reconstructions but also the cross-frame geometric alignment~\cite{xu2024scream}. 

Our discriminator follows a multi-scale PatchGAN design~\cite{esser2021taming}, capable of detecting geometric misalignments at both global and local levels. During discriminator training, we utilize a hinge loss formula to improve stability:
\begin{equation}
\mathcal{L}_{\text{con}} = \max(0, 1 - \mathbf{\Phi}(\mathbf{I}_{\text{real}})) + \max(0, 1 + \mathbf{\Phi}(\mathbf{I}_{\text{fake}})),
\end{equation}
where $\mathbf{\Phi}$ represents the discriminator network. When training the NeRF generator, the consistency loss is defined as $\mathcal{L}_{\text{con}} = -\mathbf{\Phi}(\mathbf{I}_{\text{fake}})$. In addition to the discriminative supervision, we also follow prior arts~\cite{tao2024lidar,zheng2024lidar4d,xue2024geonlf} to incorporate 2D range image supervision. The overall training objective for neural LiDAR fields combines the consistency loss, range image loss, and spectral loss. Please refer to the supplement for more details on optimization strategies.

\section{Experiment}

In this section, we first describe experimental setup for evaluation. Following this, we compare SG-NLF with previous LiDAR view synthesis methods~\cite{xue2024geonlf,zheng2024lidar4d,li2022pcgen,tao2024lidar} to demonstrate our state-of-the-art performance. Furthermore, we conduct ablation studies to expound on our specific designs.


\begin{figure*}[t]
    \centering
    \includegraphics[width=1\linewidth]{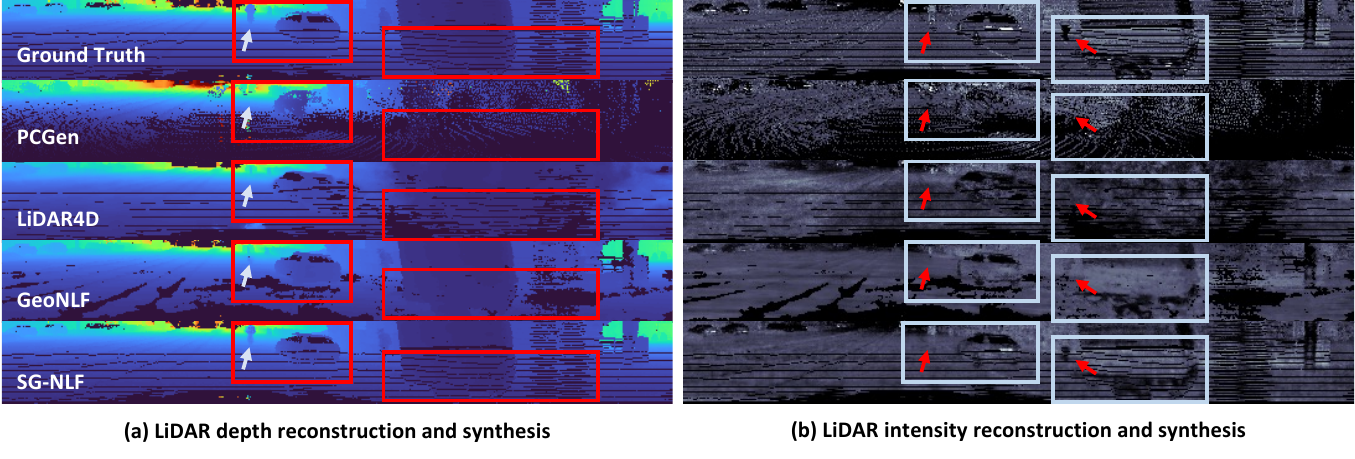}
    \vspace{-23pt}
    \caption{\textbf{Qualitative comparisons for LiDAR range depth and intensity reconstruction.} Both pose-dependent~\cite{zheng2024lidar4d,li2022pcgen} and pose-free methods~\cite{xue2024geonlf} are compared. Regions with obvious differences are highlighted in the rectangular boxes and arrows.}
    \vspace{-6pt}
    \label{fig:vis_depth_intensity}
\end{figure*}

\begin{table*}
\caption{\textbf{Comparisons on the KITTI-360~\cite{liao2022kitti} dataset with standard-frequency setting~\cite{tao2024lidar}.} We compare SG-NLF with state-of-the-art pose-dependent methods~\cite{manivasagam2020lidarsim,li2022pcgen,zheng2024lidar4d,huang2023neural,tao2024lidar} and pose-free model~\cite{xue2024geonlf}. We color
the best results as \textcolor{red}{red} and the second-best as \textcolor{orange}{orange}.}
    \vspace{-5pt}
    \centering
    \resizebox{\textwidth}{!}{
    \begin{tabular}{lccccccccccccccc}

        \hline
        \multirow{2}*{Method}  & \multicolumn{3}{c}{Point Cloud} & \multicolumn{6}{c}{Depth} & \multicolumn{5}{c}{Intensity}\\
        \cmidrule{2-3} \cmidrule{5-9} \cmidrule{11-15}
                            & CD$\downarrow$ &F-score$\uparrow$& &RMSE$\downarrow$ & MedAE$\downarrow$ & LPIPS$\downarrow$& SSIM$\uparrow$& PSNR$\uparrow$ && RMSE$\downarrow$ & MedAE$\downarrow$ & LPIPS$\downarrow$& SSIM$\uparrow$& PSNR$\uparrow$\\
        \hline
        LiDARsim~\cite{manivasagam2020lidarsim}            & 2.2249 & 0.8667 & & 6.5470  & 0.0759  & 0.2289& 0.7157  & 21.7746 && 0.1532     & 0.0506   & 0.2502   & 0.4479  & 16.3045\\
        
        PCGen~\cite{li2022pcgen}         & 0.2090   & 0.8597 & & 4.8838  & 0.1785  & 0.5210  & 0.5062  & 24.3050 & & 0.2005     & 0.0818   & 0.6100   & 0.1248  & 13.9606\\
        NKSR~\cite{huang2023neural} & 0.5780 & 0.8685 & & 4.6647  & 0.0698  & 0.2295 & 0.7052 & 22.5390 & & 0.1565     & 0.0536   & 0.2429   & 0.4200 & 16.1159\\
        LiDAR-NeRF~\cite{tao2024lidar}            & 0.0923   & 0.9226 & & 3.6801 & 0.0667 & 0.3523  & 0.6043 & 26.7663 & & 0.1557    & 0.0549     & 0.4212    & 0.2768  & 16.1683\\
        LiDAR4D~\cite{zheng2024lidar4d}           & \cellcolor{orange!40}0.0894    & \cellcolor{orange!40}0.9264  & & 3.2370  & \cellcolor{red!40}0.0507   &\cellcolor{orange!40}0.1313   & 0.7218   & 27.8840 && \cellcolor{orange!40}0.1343     & \cellcolor{orange!40}0.0404   & \cellcolor{orange!40}0.2127   & 0.4698   & \cellcolor{orange!40}17.4529\\
        GeoNLF~\cite{xue2024geonlf}           & 0.1855   & 0.9157 & & \cellcolor{orange!40}2.7144  & 0.0589  & 0.2495  & \cellcolor{orange!40}0.7797  & \cellcolor{orange!40}29.3912 && 0.1484     & 0.0613  & 0.3483  & \cellcolor{orange!40}0.3617  &16.5709\\
        SG-NLF (Ours)            & \cellcolor{red!40}0.0867  & \cellcolor{red!40}0.9322 && \cellcolor{red!40}1.8519 & \cellcolor{orange!40}0.0515 & \cellcolor{red!40}0.0864 & \cellcolor{red!40}0.9168 &\cellcolor{red!40}32.7245& & \cellcolor{red!40}0.1054    &  \cellcolor{red!40}0.0380   & \cellcolor{red!40}0.1802  & \cellcolor{red!40}0.6122 & \cellcolor{red!40}19.5468\\
        \hline
    \end{tabular}}
    \vspace{-10pt}

    \label{tab:kitti high}

\end{table*}

\subsection{Experimental Setup}

\noindent \textbf{Datasets.} We conduct experiments on two autonomous driving datasets: KITTI-360~\cite{liao2022kitti} and nuScenes~\cite{caesar2020nuScenes}. KITTI-360~\cite{liao2022kitti} is equipped with a LiDAR of 64-beam, a 26.4$^\circ$ vertical FOV, and an acquisition frequency of 10$\,$Hz. NuScenes~\cite{caesar2020nuScenes} is captured by a 32-beam LiDAR with a 40$^\circ$ vertical FOV and and an acquisition frequency of 20$\,$Hz.

\noindent \textbf{Experimental Settings.} To evaluate the reconstruction performance in low-frequency sequences, we follow GeoNLF~\cite{xue2024geonlf} to resample both datasets at $2$$\,$Hz. For KITTI-360~\cite{liao2022kitti}, we extract 24 frames from resampled sequences, retaining one sample every eight frames for testing. For nuScenes~\cite{caesar2020nuScenes}, we extract 36 frames and retain one sample every nine frames for testing. 
In addition, we conduct experiments in standard-frequency scenarios following LiDAR-NeRF~\cite{tao2024lidar}, extracting 64 consecutive frames and four equidistant test samples from KITTI-360~\cite{liao2022kitti}. LiDAR poses are perturbed with additive noise using a standard deviation of $20^\circ$ in rotation and $3m$ in translation~\cite{xue2024geonlf}.

\noindent \textbf{Evaluation Metrics.} We evaluate the reconstruction accuracy of LiDAR point clouds using the Chamfer Distance (CD) and the F-score value (error threshold of $5cm$)~\cite{xue2024geonlf,tao2024lidar,zheng2024lidar4d,yu2025stgc}. For novel view synthesis, we report depth and intensity errors of the projected range images using Root Mean Square Error (RMSE) and Median Absolute Error (MedAE), while overall variance is measured with LPIPS~\cite{zhang2018unreasonable}, SSIM~\cite{wang2004image}, and PSNR. For pose evaluation, we follow GeoNLF~\cite{xue2024geonlf} to use standard odometry metrics, including Absolute Trajectory Error (ATE) and Relative Pose Error (RPE$_r$ for rotation and RPE$_t$ for translation).

\noindent\textbf{Implementation Details.} Consistent with prior arts~\cite{tao2024lidar,zheng2024lidar4d,xue2024geonlf,yu2025stgc}, the entire point cloud scene is scaled within the unit cube space and $768$ points are uniformly sampled along each laser beam. Our SG-NLF is implemented on PyTorch~\cite{paszke2019pytorch} and optimized with the Adam~\cite{kingma2014adam} optimizer. Each scene is trained for $60$k iterations with a batch size of $4,096$ rays. The initial learning rate is $0.01$ and follows a linear power decay schedule, as in previous work~\cite{xue2024geonlf}.  We construct hash grids based on tiny-cuda-nn. All experiments are conducted on a single GeForce NVIDIA RTX 4090 GPU. Refer to the supplement for more details.

\subsection{Comparisons in LiDAR View Synthesis}
\noindent\textbf{Comparisons in Low-frequency Scenarios.} We compare our method with state-of-the-art approaches for LiDAR novel view synthesis, including pose-dependent~\cite{manivasagam2020lidarsim,li2022pcgen,zheng2024lidar4d} and pose-free methods~\cite{lin2021barf,heo2023robust,xue2024geonlf,qin2022geometric}. For approaches not originally designed for LiDAR NVS~\cite{lin2021barf,heo2023robust,qin2022geometric}, we follow GeoNLF~\cite{xue2024geonlf} and adopt LiDAR-NeRF~\cite{tao2024lidar} for reconstruction.  For pose-dependent methods, we employ ground-truth poses for synthesis. For pose-free approaches, we follow prior arts~\cite{xue2024geonlf,wang2021nerf} to obtain test view poses for rendering.

Table~\ref{tab:kitti} demonstrates our state-of-the-art performance on KITTI-360~\cite{liao2022kitti} with a low-frequency setting~\cite{xue2024geonlf}. Our SG-NLF significantly surpasses all these methods in point cloud reconstruction and range image synthesis. Compared to the second-best pose-free method GeoNLF~\cite{xue2024geonlf}, SG-NLF reduces CD by 28.3\%, while improving depth PSNR and intensity PSNR by 13.6\% and 16.2\% respectively.
Notably, even when LiDAR4D~\cite{zheng2024lidar4d} utilizes ground-truth poses for reconstruction and synthesis, pose-free SG-NLF still outperforms it by over 38.5\%, 37.5\%, and 25.4\% for CD, depth RMSE, and intensity RMSE, showing the efficacy of proposed architectures in challenging low-frequency scenarios. 

Similarly, in Table~\ref{tab:nus}, compared to the leading method~\cite{xue2024geonlf}, SG-NLF reduces reconstruction errors by over 37.5\%, 47.2\%, 73.7\%, 20.9\%, and 54.9\% for CD, depth RMSE, depth LPIPS, intensity RMSE, and intensity LPIPS on low-frequency nuScenes~\cite{caesar2020nuScenes,xue2024geonlf}. This superior performance 
demonstrates the effectiveness of our spectral-geometric design for sparse LiDAR data. 

\noindent\textbf{Comparisons in Standard-frequency Scenarios.} Following prior arts~\cite{tao2024lidar,zheng2024lidar4d}, we evaluate SG-NLF on the standard-frequency KITTI-360 dataset~\cite{liao2022kitti}. As shown in Table~\ref{tab:kitti high}, compared to state-of-the-art pose-dependent methods~\cite{manivasagam2020lidarsim,li2022pcgen,zheng2024lidar4d,huang2023neural,tao2024lidar}, our pose-free SG-NLF still achieves more accurate reconstruction and synthesis. While the pose-free GeoNLF~\cite{xue2024geonlf} underperforms pose-dependent LiDAR4D~\cite{zheng2024lidar4d} on most metrics, SG-NLF improves depth PSNR and intensity PSNR by over 17.4\% and 11.9\% compared to LiDAR4D.  This demonstrates the generalization of our framework in complex real-world scenes. The aforementioned results showcase our effectiveness in both low-frequency and standard-frequency scenarios.

\begin{figure*}[t]
    \centering
    \includegraphics[trim=0cm 0cm 0cm 0cm, clip, width=1\linewidth]{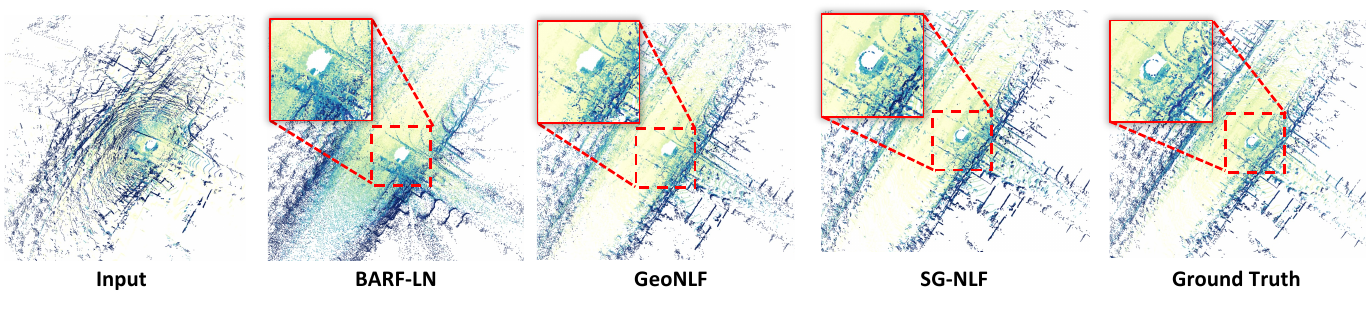}
    \vspace{-15pt}
    \caption{\textbf{Visual comparisons for pose estimation.} We transform input sequences into a unified scene using estimated global poses and color-code the composite scene by height. We compare with pose-free methods~\cite{lin2021barf,xue2024geonlf,tao2024lidar}. Best view zoomed in on-screen for details.}
    \vspace{-10pt}
    \label{fig:pose}
\end{figure*}

\noindent\textbf{Qualitative Comparisons.} Fig.~\ref{fig:vis_depth_intensity} visualizes depth and intensity reconstruction results in challenging low-frequency scenarios from KITTI-360~\cite{liao2022kitti}. We compare our method with both pose-dependent~\cite{zheng2024lidar4d,li2022pcgen} and pose-free~\cite{xue2024geonlf} models. As highlighted by the rectangles and arrows, LiDAR4D~\cite{zheng2024lidar4d} and GeoNLF~\cite{xue2024geonlf} produce blurry results with missing structural details. In contrast, our SG-NLF generates accurate depth and intensity maps that depict smooth and complete structures while preserving clear boundaries (\textit{e.g.}, pedestrians and vehicles). Furthermore, the black horizontal lines in these maps represent the predicted raydrop mask. While competing methods~\cite{zheng2024lidar4d,li2022pcgen,xue2024geonlf} exhibit substantial erroneous predictions, our results align closely with the ground truth. These improvements can be attributed to our effective spectral-geometric hybrid representation and cross-frame consistency strategy. Please refer to the supplement for more quantitative and visual experimental results.


\subsection{Comparisons in Pose Estimation} 
\noindent\textbf{Quantitative Comparisons.} We follow prior art~\cite{xue2024geonlf} to compare SG-NLF with both pairwise registration baselines~\cite{besl1992method,qin2022geometric,lu2021hregnet,choi2015robust,wang2023robust} and pose-free NeRF methods~\cite{lin2021barf,heo2023robust,xue2024geonlf,tao2024lidar}.  For pairwise methods, we perform registration between adjacent frames in an odometry-like manner. All estimated trajectories are aligned with the ground truth using a Sim(3) transform of known scale~\cite{xue2024geonlf}. As shown in Table~\ref{tab:pose}, SG-NLF outperforms both pairwise registration and pose-free NeRF approaches on the KITTI-360~\cite{liao2022kitti,xue2024geonlf} and nuScenes~\cite{caesar2020nuScenes,xue2024geonlf} datasets. Specifically, SG-NLF reduces ATE by over 56.4\% and 68.8\% on KITTI-360 and nuScenes compared to the second-best GeoNLF~\cite{xue2024geonlf}.

\noindent\textbf{Qualitative Comparisons.} To visually evaluate pose accuracy, we transform point cloud sequences into a unified scene using estimated global poses. We color-code the composite scene by height (a spectrum from yellow to purple) in Fig.~\ref{fig:pose}. Accurate poses produce coherent color gradations (see ground truth), while pose errors result in misalignment and color fragmentation (see input sequences). BARF-LN~\cite{lin2021barf,tao2024lidar} retains significant noise in road areas, indicating inaccurate estimation. 
Conversely, our SG-NLF produces accurately aligned and geometrically consistent point clouds, achieving quality closest to the ground truth.

\begin{table}
    \caption{\textbf {Pose estimation comparisons on KITTI-
360~\cite{liao2022kitti,xue2024geonlf} and nuScenes~\cite{caesar2020nuScenes,xue2024geonlf}.} We compare with registration baselines~\cite{besl1992method,qin2022geometric,lu2021hregnet,choi2015robust,wang2023robust} and pose-free NeRF methods~\cite{lin2021barf,heo2023robust,xue2024geonlf,tao2024lidar}.}
\vspace{-3pt}
    \centering
    \resizebox{\linewidth}{!}{
    \begin{tabular}{lccccccc}
        \hline
        \multirow{3}*{Method}  & \multicolumn{4}{c}{KITTI-360}          & \multicolumn{3}{c}{nuScenes}      \\
        \cmidrule{2-4} \cmidrule{6-8}
        & ATE & RPE$_r$ & RPE$_t$ && ATE & RPE$_r$ & RPE$_t$ \\
        &(m)$\downarrow$&(deg)$\downarrow$ &(cm)$\downarrow$ &&(m)$\downarrow$&(deg)$\downarrow$ &(cm)$\downarrow$\\
        \hline
        ICP~\cite{besl1992method}                 & 1.894 & 1.019 & 30.383 & & 1.131 & 0.647 & 15.410 \\
        MICP~\cite{choi2015robust}               & 1.483 & 1.419 & 35.584 & & 2.519 & 1.101 & 38.840\\
        HRegNet~\cite{lu2021hregnet}             & 7.423 & 9.083 & 290.160& & 7.815 & 2.173 & 120.913 \\
        SGHR~\cite{wang2023robust}                  & 2.539 & 0.906 & 95.576 & & 9.557 & 0.699 & 30.383\\
        BARF-LN~\cite{lin2021barf,tao2024lidar}                   & 2.763 & 2.203 & 199.740& & 5.244 & 0.819  & 210.331  \\
        HASH-LN~\cite{heo2023robust,tao2024lidar}              & 2.666 & 2.171 & 196.791& & 4.151 & 0.832 & 180.282 \\
        GeoTrans~\cite{qin2022geometric}                  & 0.246 & 0.213  & 6.081 & & 0.892 & 0.363  & 16.097 \\
        GeoNLF~\cite{xue2024geonlf}              & \cellcolor{orange!40}0.170 & \cellcolor{orange!40}0.205  & \cellcolor{orange!40}5.449  && \cellcolor{orange!40}0.228  & 0.\cellcolor{red!40}103  & \cellcolor{orange!40}7.058  \\

        SG-NLF(Ours)                 & \cellcolor{red!40}0.074 & \cellcolor{red!40}0.127 & \cellcolor{red!40} 2.003 & & \cellcolor{red!40}0.071 & \cellcolor{orange!40}0.158 & \cellcolor{red!40}4.096\\
        \hline
    \end{tabular}}
    \vspace{-5pt}
    \label{tab:pose}
\end{table}

\begin{table}
\caption{\textbf{Hybrid Representation Analysis.}  Ablation study of geometric encodings (GE) and spectral embeddings (SE) on nuScenes~\cite{caesar2020nuScenes}, with comparisons to GeoNLF~\cite{xue2024geonlf}.}
\vspace{-15pt}
\begin{center}
\resizebox{\linewidth}{!}{
\begin{tabular}{lcccccc}
\toprule
 \multirow{2}*{Method} & \multirow{2}*{GE} & \multirow{2}*{SE}& Point Cloud &  Depth& Intensity  \\
 & & &CD$\downarrow$   & PSNR$\uparrow$& PSNR$\uparrow$ \\
\midrule
GeoNLF~\cite{xue2024geonlf}  & $\usym{2714}$&$\usym{2717}$&0.241&22.947&28.608 \\
SG-NLF($w/o$ GE)& $\usym{2717}$&$\usym{2714}$ & 0.181& 26.853 & 29.032  \\
SG-NLF(Ours) & $\usym{2714}$&$\usym{2714}$   & $\bold{0.155}$& $\bold{28.409}$ & $\bold{30.499}$  \\
\bottomrule
\end{tabular}}
\end{center}
\vspace{-10pt}
\label{tab:ablation_hybrid}
\end{table}

\subsection{Ablation Studies}

\begin{figure}
    \centering
    \includegraphics[width=0.9\linewidth]{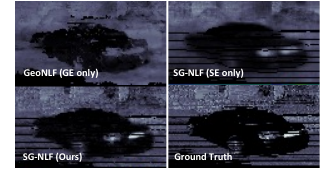}
    \vspace{-7pt}
    \caption{\textbf{Qualitative comparison for hybrid representation. }}
    \vspace{-9pt}
    \label{fig:ablation}
\end{figure}

\noindent\textbf{Hybrid Representation Analysis.} We presents an ablation study of our spectral embeddings with comparisons to GeoNLF~\cite{xue2024geonlf} in Table~\ref{tab:ablation_hybrid}. While GeoNLF relies on geometric interpolation for rendering, our SG-NLF ($w/o$ GE) reconstructs more smooth and complete geometry through spectral embeddings (shown in Fig.~\ref{fig:ablation}). However, it tends to miss several high-frequency details. The hybrid representation can achieve high-quality reconstruction both quantitatively and qualitatively by integrating low-frequency spectral embeddings with high-frequency geometric encodings.

\noindent\textbf{Efficacy of SG-NLF Components.} In Table~\ref{tab:ablation_component}, we ablate on our three core components: hybrid spectral-geometric representation (HR), global pose optimization (GP), and cross-frame consistency (CFC). We adopt the same baseline as GeoNLF~\cite{xue2024geonlf}, \textit{i.e.}, a geometric encoding model without pose estimation. For the ablation of hybrid spectral-geometric representation, we retain only geometric encodings for feature extraction, resulting in decreased pose accuracy and reconstruction quality. Comparing the first and third rows, even without pose optimization, incorporating cross-frame consistency supervision effectively regularizes training and mitigates performance degradation. Conversely, as shown in the fourth row, removing CFC leads to cross-frame inconsistency and performance degradation.
\begin{table}

\caption{\textbf{Efficacy of SG-NLF Components.} Ablation on the hybrid spectral-geometric representation (HR), global pose optimization (GP), and cross-frame consistency (CFC) using nuScenes dataset~\cite{caesar2020nuScenes}. We adopt the same baseline as GeoNLF~\cite{xue2024geonlf}.}
\vspace{-12pt}
\begin{center}
\resizebox{\linewidth}{!}{
\begin{tabular}{lccccccc}
\toprule
 \multirow{2}*{Method} & \multirow{2}*{HR} & \multirow{2}*{GP}& \multirow{2}*{CFC} & Point Cloud &  Depth& Intensity   & Pose\\
 & & & &CD$\downarrow$   & PSNR$\uparrow$& PSNR$\uparrow$ & ATE(m)$\downarrow$\\
\midrule
Baseline & $\usym{2717}$& $\usym{2717}$& $\usym{2717}$ &0.618&21.321&25.856&1.328\\
$w/o$ HR & $\usym{2717}$& $\usym{2714}$& $\usym{2714}$& 0.217& 25.103 & 28.425&0.204\\  
$w/o$ GP& $\usym{2714}$&$\usym{2717}$ & $\usym{2714}$& 0.463& 23.942 & 27.549 &0.798\\  
$w/o$ CFC& $\usym{2714}$&$\usym{2714}$ & $\usym{2717}$& 0.182& 26.597 & 29.303&0.076\\  
Ours & $\usym{2714}$&$\usym{2714}$ &$\usym{2714}$  & $\bold{0.155}$& $\bold{28.409}$ & $\bold{30.499}$& $\bold{0.071}$  \\
\bottomrule
\end{tabular}}
\end{center}
\vspace{-25pt}
\label{tab:ablation_component}
\end{table}
\section{Conclusion}
In this paper, we propose SG-NLF, a spectral-geometric neural fields for pose-free LiDAR view synthesis. 
Different from previous methods that rely on geometric interpolation, our approach introduces a hybrid representation that integrates spectral embeddings to reconstruct smooth and continuous geometry. We also design a confidence-aware graph optimizer for robust global pose estimation and an adversarial learning strategy to enforce cross-frame consistency. Comprehensive experiments demonstrate the effectiveness of the proposed SG-NLF framework. Our work can provide a novel perspective on the accurate LiDAR view synthesis. 

\noindent \textbf{Limitations.} We currently provide one effective implementation of the SG-NLF. In future work, more techniques can be explored for different application scenarios.


{
    \small
    
    \bibliographystyle{ieeenat_fullname}
    \bibliography{main}
}

\end{document}